\documentclass{article}

\usepackage{tabularx}
\usepackage[utf8]{inputenc} % allow utf-8 input
\usepackage[T1]{fontenc}    % use 8-bit T1 fonts
\usepackage{hyperref}       % hyperlinks
\usepackage{url}            % simple URL typesetting
\usepackage{booktabs}       % professional-quality tables
\usepackage{amsfonts}       % blackboard math symbols
\usepackage{nicefrac}       % compact symbols for 1/2, etc.
\usepackage{microtype}      % microtypography
\usepackage{xcolor}         % colors
\usepackage{graphicx}
\usepackage{array}     % 解决 \arraybackslash 
\usepackage{multirow}  % 解决 \multirow 
\usepackage{booktabs}  % 确保 \toprule、\midrule
\usepackage{amsmath} 
\usepackage{natbib}
\usepackage{algorithm}
\usepackage{algpseudocode}
\usepackage{subfigure}
\usepackage{amssymb}
\usepackage{amsthm}
\newtheorem{assumption}{Assumption}

\newtheorem{proposition}{Proposition}

\title{ViTNF: Leveraging Neural Fields to Boost Vision Transformers in Generalized Category Discovery }

\author{
	  Jiayi Su
		     \\
	 School of Mathematics and Information
	 Science\\
	Guangxi University\\
	Nanning, China 530004 \\
	\texttt{2306301032@st.gxu.edu.cn} \\
	 Dequan Jin\thanks{Corresponding author} \\
	 School of Mathematics and Information
	 Science\\
	 Guangxi University\\
	 Nanning, China 530004 \\
	 \texttt{dqjin@gxu.edu.cn} \\
	 Shihui Ying
	 \\
	 Shanghai Institute of Applied Mathematics and Mechanics\\
	 School of Mechanics and Engineering Science\\
	 Shanghai University\\
	 Shanghai, China 200072 \\
	 \texttt{shying@shu.edu.cn} \\	 
		}

\begin{document}

	\maketitle

	\begin{abstract}
				Generalized category discovery (GCD) is a highly popular task in open-world recognition, aiming to identify unknown class samples using known class data. By leveraging pre-training, meta-training, and fine-tuning, the vision transformer (ViT) achieves excellent few-shot learning capabilities in GCD tasks. However, most improvements on ViT focus on its feature extractor module, including patch and position embedding parts and encoder, but seldom discuss improving its classifier module, the MLP Head. The MLP head is a feedforward network trained synchronously with the entire network in the same error back-propagation process, increasing the training cost and difficulty without fully leveraging the power of the feature extractor. For these issues, this paper proposes a new architecture by replacing the MLP head with a neural field-based classifier. We first present a new static neural field function to describe the activity distribution of the neural field and build an efficient neural field-based (NF) classifier with it. It stores the feature information of support samples by its elementary field, the known categories by its high-level field, and the category information of support samples by its cross-field connections. We replace the MLP head with the proposed NF classifier, resulting in a novel architecture ViTNF, and simplify the three-stage training mode by pre-training the feature extractor on source tasks and training the NF classifier with support samples in meta-testing separately, significantly reducing ViT's demand for training samples and the difficulty of model training. To enhance the model's capability in identifying new categories, we provide an effective algorithm to determine the lateral interaction scale of the elementary field. Experimental results demonstrate that our model surpasses existing state-of-the-art methods on CIFAR-100, ImageNet-100, CUB-200, and Standard Cars, achieving dramatic accuracy improvements of 19\% and 16\% in new and all classes, respectively, indicating a notable advantage in GCD.
	\end{abstract}
	
	\section{Introduction}\label{introduction}
	%The image recognition of machines simulates how the human brain recognizes an image. However, we observe that humans would not limit themselves to their existing knowledge but rather develop curiosity about unfamiliar objects in daily life, leading to further learning\cite{CDL1}. For example, imagine we have already recognized "apples" and "pears." Then, a new object called "banana" appears on our table, and we haven't seen it before. At this point, we begin to observe this totally new object. We may notice that, in terms of shape, this unknown object has no apparent connection to the "apples" or "pears" we have previously learned before. However, in terms of color, it may share similarities with certain types of "apples" and "pears." Then, we try to consider whether this object belongs to a new category. After further analysis, we would determine that it is a new category. Although we might not yet know its name to be "banana," we have already identified it as something distinct from "apples" and "pears" and categorized it as an entirely new class.This is an entire learning process for humans. Therefore, we need to consider the methods provide machines with generalized category discovery (GCD) capability to simulate this process of reasoning and learning in real open-wolrd scenarios\cite{GCD}.\par
	In image classification, we hope machines can recognize images as humans do\cite{simulate_human}. Currently, supervised learning algorithms can identify the known categories in the training samples, and unsupervised algorithms can discover underlying clusters of unlabeled samples. However, in real-world open-world tasks, there may be such scenarios: we need to classify unlabeled samples from some new, unseen categories based on labeled data of known categories. Such problems are called generalized category discovery (GCD)\cite{GCD}. GCD is not difficult for humans\cite{CDL1}. For example, suppose we have recognized apples and pears after training with their labeled samples. When a new fruit sample appears, we can identify whether it belongs to apples, pears, or a new category based on its visual feature similarities with the two known types of fruits. However, GCD is a real challenge for machines because they have no information about new categories and need to rely entirely on the information of known categories to complete the recognition. At the same time, most GCD tasks are also in the few-shot learning (FSL) scenarios, where the labeled samples of known categories are very few. It requires the learning models for GCD tasks to have good FSL capabilities.
	
\begin{figure*}[htb]
		\centering
		\subfigure[ViT]{
			\includegraphics[width=0.5\textwidth]{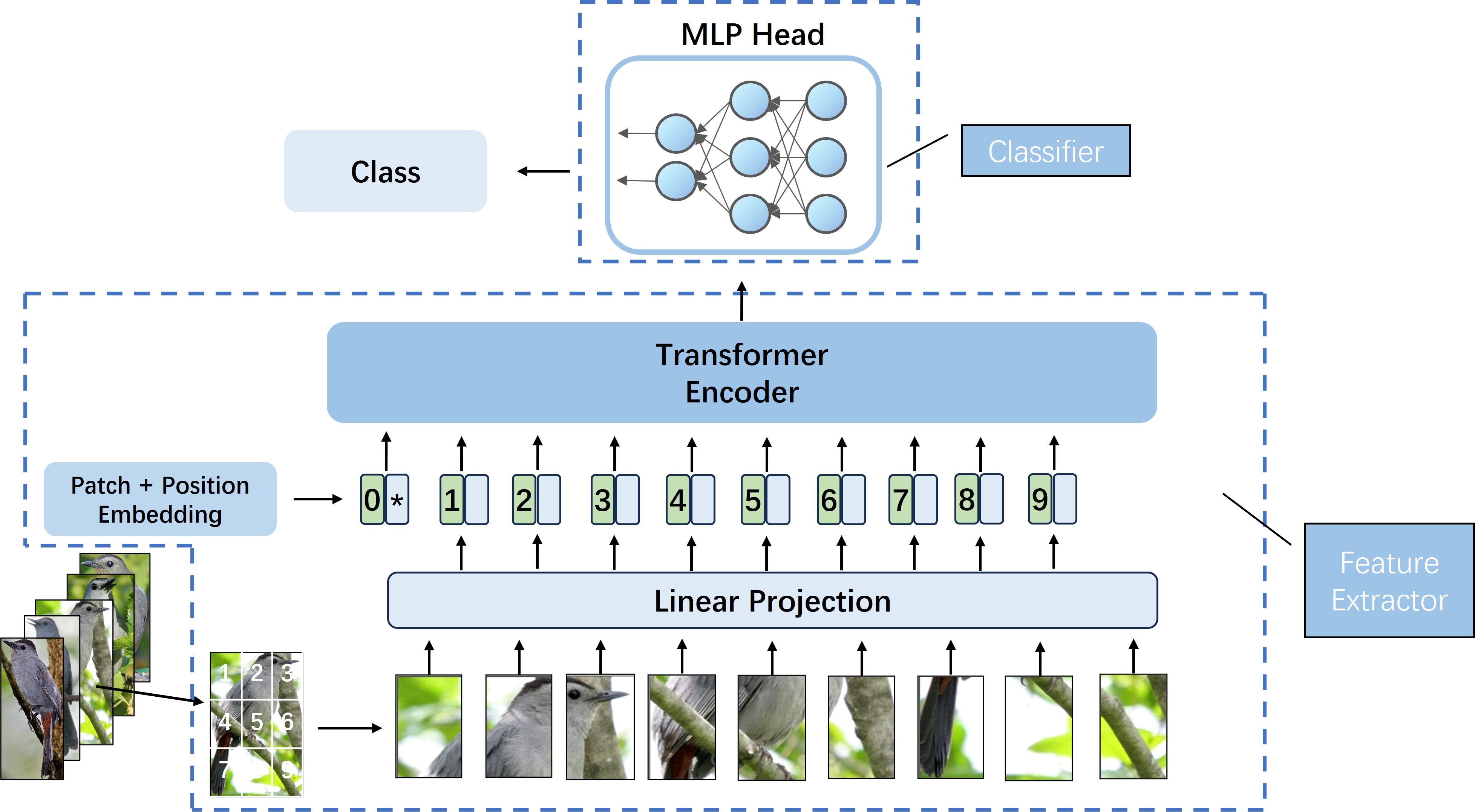}
		}
		\subfigure[ViTNF]{
			\includegraphics[width=0.4\textwidth]{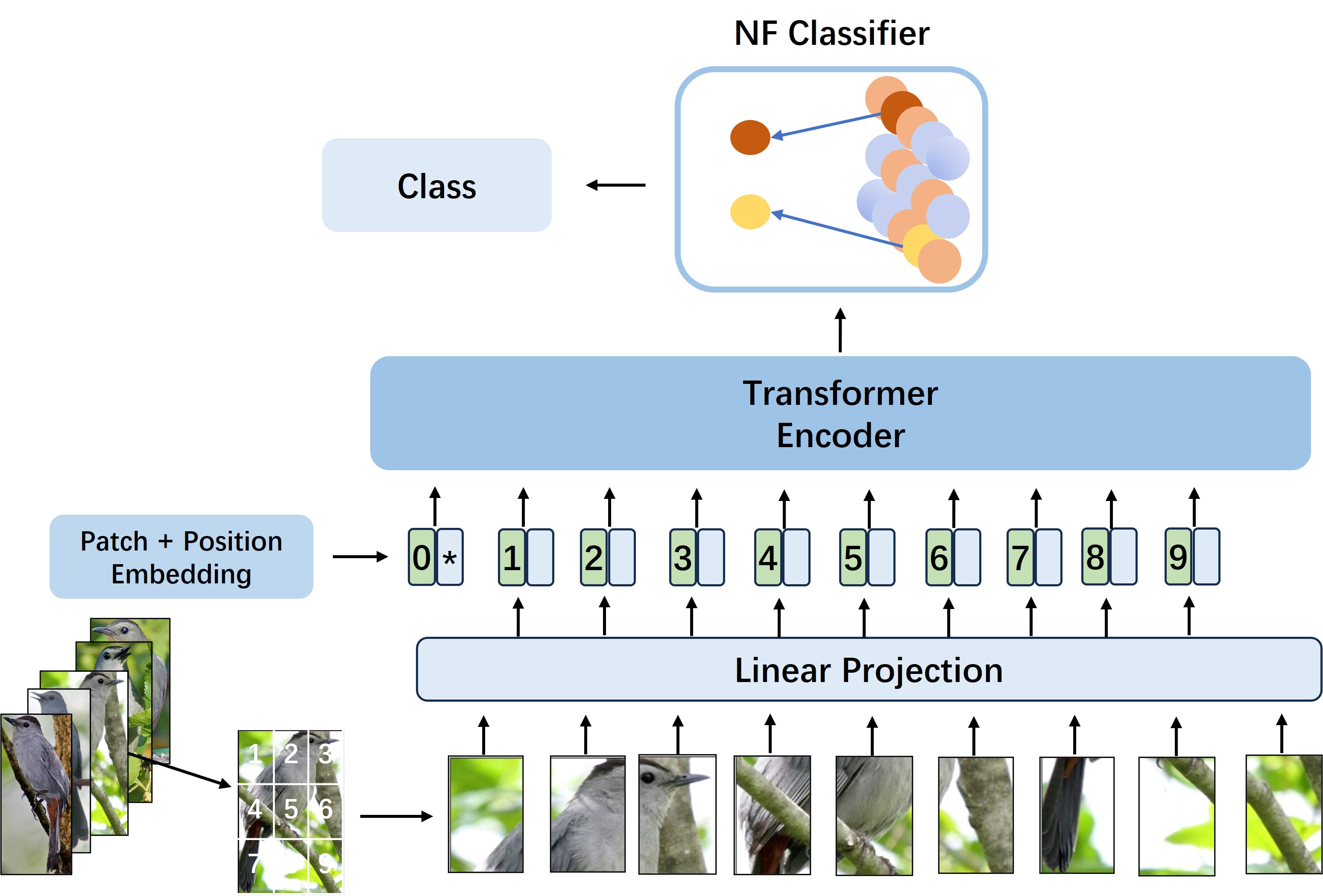}
		}
		\caption{The structures of (a) ViT and (b) ViTNF. }\label{fig:structure}
	\end{figure*}

	Few-shot learning refers to the process in which a machine learns and recognizes using only a limited number of images and their corresponding labels. With the rapid development of FSL technology, GCD has also made significant progress in open-world recognition. By applying clustering\cite{km}, feature learning and extraction, and category matching and selection\cite{Brent}, FSL models are capable of dealing with GCD tasks. GCD requires the learning models to possess powerful representation learning capability. Since vision transformer (ViT) possesses dramatic feature representation capabilities\cite{VIT}, it is widely used as the backbone of few-shot learning models, providing these models with excellent global information capture ability and strong generalization ability. The overall network structure of ViT can be divided into two modules: the feature extractor and the classifier, as shown in Figure \ref{fig:structure}a. The feature extractor module consists of the linear projection, the patch+position embedding, and the transformer encoder. They transform image samples into their feature vectors. The classifier module is the MLP Head. It is a multi-layer perceptron for classifying samples according to their feature vectors. 
	
	%In traditional machine learning tasks, machines need significant images with corresponding labels as a training set, allowing them to learn from the dataset. However, this approach is often difficult to implement in real-world scenarios, such as medical image recognition. To address this issue, researchers have proposed few-shot learning methods, which enable machines to learn and recognize images with limited image and label resources\citep{FSL1,FSL2,FSL3,FSL4}. Alexey Dosovitskiy et al. found through research that the Transformer model can be directly used for image recognition. Building upon the existing ViT model, we replaced its original image classification module with a neural field-based image recognition model, enabling it to conduct few-shot learning tasks.\cite{VIT} Current neural networks primarily simulate the activation states of individual neurons and their associated neural networks\citep{Neu1,Neu2,Neu3}. The connection weights between different neurons are adjusted through algorithms, forming structures that can mimic a small portion of the nervous system\citep{Neu4,Neu5}. Unlike other models, our neural field models do not focus solely on individual neurons and their connections. Instead, they observe the behavioral states of neuronal populations within specific regions.\par
	
\begin{figure*}[htb]
		\centering
		\subfigure[ViT]{
			\includegraphics[width=0.3\textwidth]{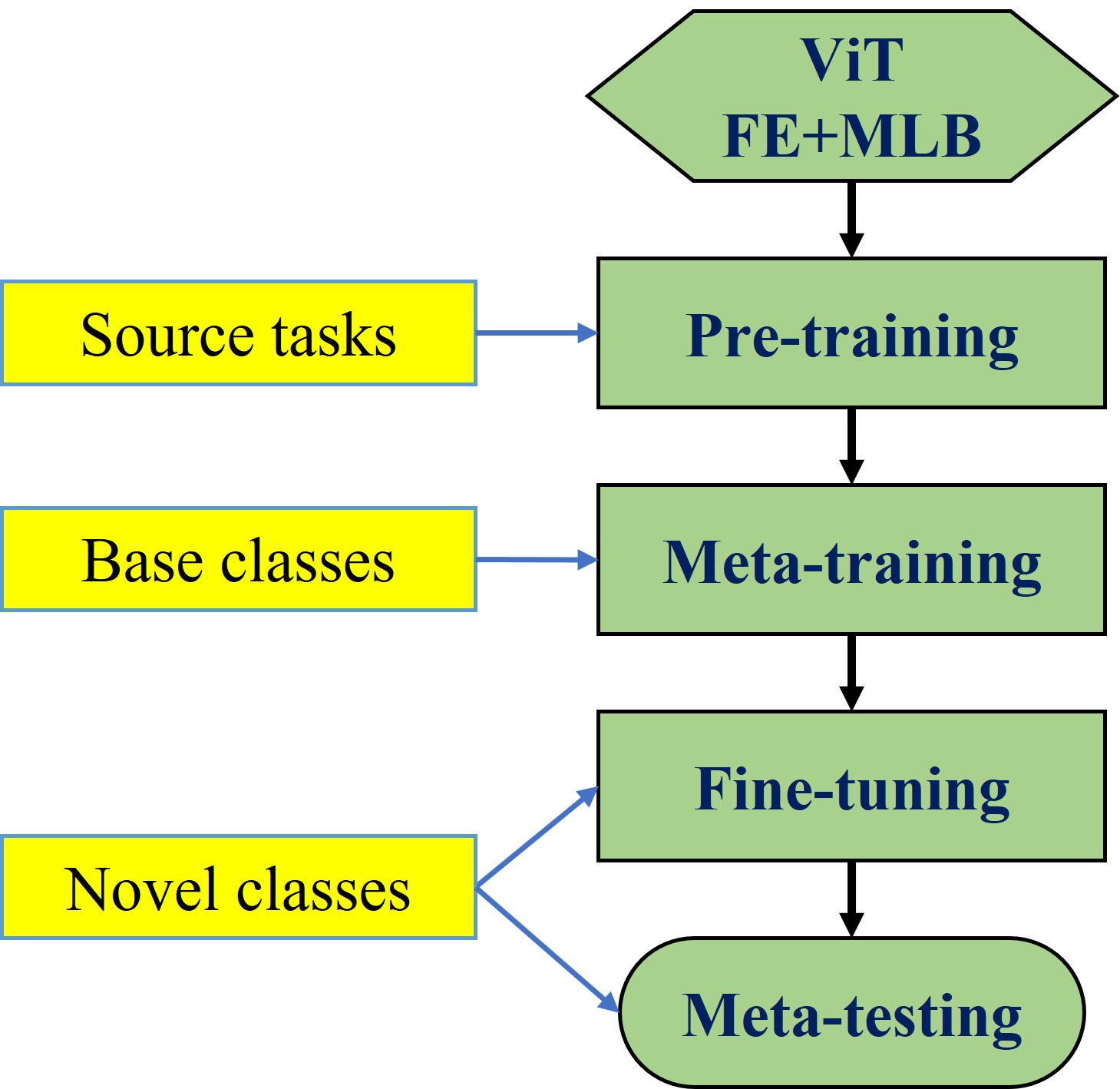}
		}
		\subfigure[ViTNF]{
			\includegraphics[width=0.6\textwidth]{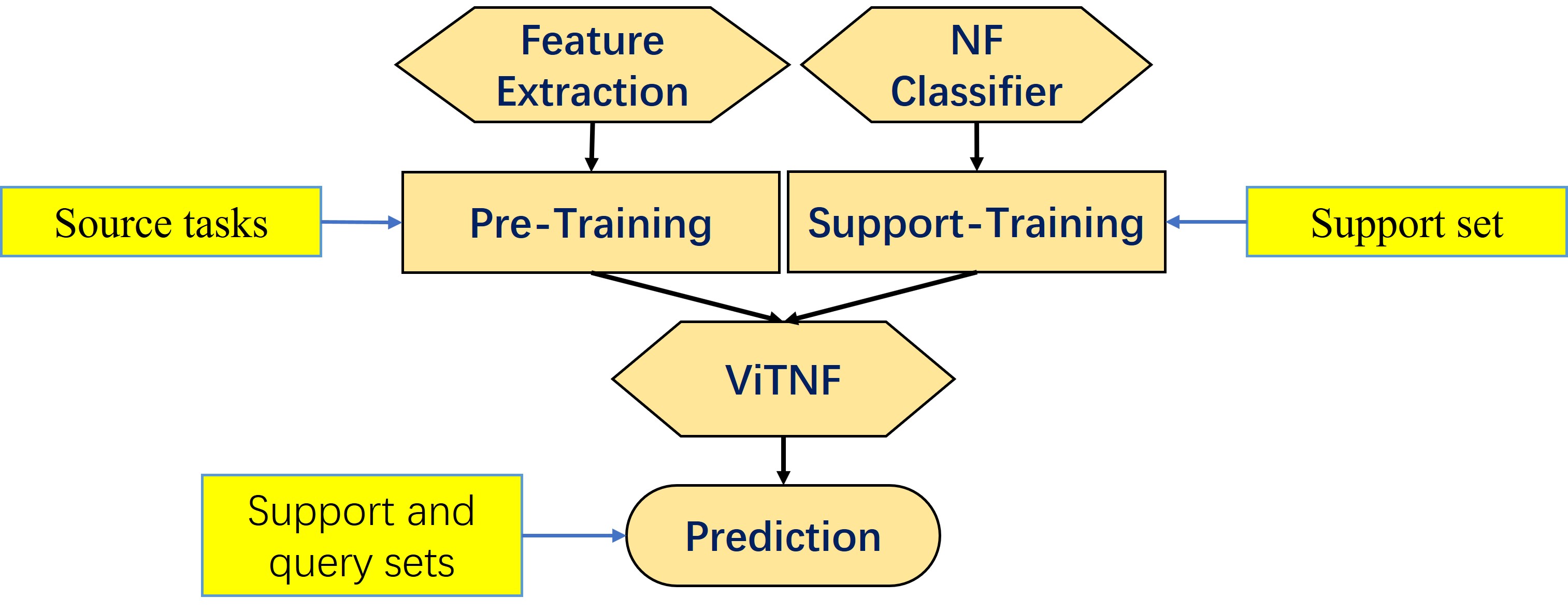}
		}
		\caption{The training processes of (a) ViT and (b) ViTNF. }\label{fig:tp}
	\end{figure*}

	To enhance its FSL performance, we can leverage a three-stage training strategy to train a ViT: pre-training the network in source tasks and meta-training and fine-tuning it in a target task, as shown in Figure \ref{fig:tp}a. These training stages provide ViT with excellent FSL performance in the meta-testing stage, where the network utilizes the knowledge obtained in these training stages to infer the novel few-shot task, making it a popular backbone in many FSL methods. However, they also lead to some issues in FSL. For instance, the entire network is trained simultaneously through the three-stage training. It seems simple to design the training strategy, but the training processes of the feature extractor and the classifier have different requirements. The feature extractor relies more on pre-training. We can achieve an excellent feature extractor by pre-training it with a large sample. Nonetheless, the training of MLP relies more on the support samples in the meta-testing and benefits very little from pre-training. These differences make the simultaneous training less efficient and cannot sufficiently leverage the sample information. Moreover, MLP is essentially a feedforward neural network. Its training relies on the error back-propagation algorithm and has high requirements for the sample size. Since pre-training and meta-training have few effects on improving the classification performance of MLP, and the samples available for meta-testing are limited, it makes it difficult to fully leverage the excellent performance of the feature extractor, thereby restricting the overall performance of ViT. Moreover, MLP is a typical supervised learning network. It is not for discovering new categories. The neurons in its output layer indicate the learned categories. In a GCD task, we must add new neurons to the output layer when new categories are detected. This operation may increase the training difficulty and decrease the network performance,  perhaps inducing catastrophic forgetting in learning new categories\cite{simGCD}. If we design a new classifier suitable for identifying new categories, we can replace the MLP head with it in ViT to enhance the training efficiency and GCD capability.  
	
	In current research on neural networks, neural fields are a type of neural model frequently used for small sample learning. The neural field model was first proposed in the 1970s and was used to describe the dynamical spatio-temporal average activity behavior of neurons in the cerebral cortex. Unlike the traditional feedforward neural networks, having multiple layers composed of some isolated neurons with adaptable connection weights determined by the error back-propagation (BP) algorithms, a neural field describes neuronal activation with a spatially continuous field. The connection weights between its neurons are fixed, determined by their distance. The neural field represents the input pattern by its activation distribution. The neural activity in neural fields can quickly adapt to their input and achieve a one-to-one correspondence with the spatial orientation of the visual field. Therefore, they have a significantly small sample learning potential and are often used to describe the neuronal activation of working and short-term memory. In engineering, neural fields are also used in fields such as robot control, pattern recognition, small sample learning, and unsupervised learning, demonstrating excellent rapid learning capabilities and having the potential to construct efficient few-sample classifiers.
	
	%However, excessive dependence on known categories may lead to overfitting, making it difficult to correctly identify new categories, which causes an imbalance in accuracy between old and new categories. \par
	%Incremental learning is a machine-learning strategy designed to enable a model to continuously learn new tasks or classes of data without forgetting the knowledge known before. Unlike traditional training methods, incremental learning does not rely on accessing all data simultaneously; instead, it gradually introduces new data or classes\citep{IL1,IL2,IL3}. The goal of GCD is to identify samples from the unlabeled data that belong to entirely new categories different from any known labels and learn from them. Therefore, when modifying our model, we incorporated the idea of incremental learning and introduced an additional step in the testing phase specific for the selection and learning of new categories.\par
	This paper presents an effective GCD method. We construct a highly efficient classifier based on neural fields, namely the neural field-based head (NF head), and replace the MLP head in ViT with it. To achieve better computing efficiency, we propose a static neural field function by analyzing the steady state of the dynamical neural field. After that, we construct an NF head with two static neural fields. The NF head consists of two static neural fields. One is the elementary field, corresponding to a feature space. Its input is the feature vector obtained by the feature extractor. The other is the high-level neural field, whose neurons correspond to the sample categories. During training with support samples in meta-testing, we connect the elementary neurons corresponding to the feature vectors of the support samples to the advanced neurons corresponding to their categories, thereby memorizing their positional information and utilizing the lateral interactions between the primary neurons to achieve the generalization capability in the prediction stage. By embedding the NF head into ViT and replacing the original MLP head, the resulting ViT+NF head (ViTNF) model achieves state-of-the-art few-shot classification performance by only pre-training the feature extractor and support-training the NF head. To validate its FSL performance, we evaluate the proposed model on the  CIFAR-10, CIFAR-100, ImageNet-100, and CUB-200 and Stanford Cars datasets in semantic shift benchmark (SSB) for GCD in 5-way 1-shot and 5-way 5-shot tasks. In summary, we list our main contributions as follows:
	\begin{enumerate}
		\item We propose a static neural field-based (NF) classifier to replace the MLP head of ViT. The NF classifier can quickly learn from the support samples without BP algorithms, significantly improving the entire network's training efficiency.
		\item We propose effective learning strategies to enhance the performance of the NF classifier, providing it with excellent FSL and GCD capabilities and accuracy.
		\item We simplify the original three-stage training mode by pre-training the feature extractor on source tasks and training the classifier with support samples in the meta-testing separately, significantly reducing ViT's demand for training samples and the difficulty of training.
		\item Extensive experiments demonstrate that the original ViT can achieve superior GCD classification accuracy by replacing its MLP head with the proposed NF classifier without meta-training or fine-tuning, outperforming state-of-the-art methods in both old and new classes.
	\end{enumerate}
We organize this paper as follows. We present the related works in Section \ref{sec:relatedworks}, briefly introduce few-shot learning and neural field equations in Section \ref{sec:preli}, then propose the architecture of the NF-based classifier and the learning strategies in Section \ref{sec:method}. We provide some extensive experiments on real-world benchmark datasets and ablation studies in Section \ref{Experiment}. Finally, the conclusion is in Section \ref{conclusion}.
	
	\section{Related Works}\label{sec:relatedworks}
	\subsection{Generalized category discovery}
	In recent years, studies on the open-world problem and GCD have emerged. The open-world learning ORCA is an end-to-end open-world deep learning method\cite{orca}. It introduces an uncertainty-adaptive boundary mechanism to avoid bias toward known classes due to the faster learning of discriminative features for seen classes. OpenLDN is an open-world SSL method with the core idea of detecting new classes through pairwise similarity loss\cite{openLDN} by recognizing samples from known classes and detecting new classes in unlabeled data simultaneously. 
	
Some methods acquire information about new categories from the unlabeled samples by unsupervised or semi-supervised learning. In 2022, S. Vaze et al. proposed the concept of generalized category discovery (GCD) and used Hungarian and Brent algorithms to estimate the number of unknown categories for clustering\cite{GCD}. $\mu$GCD method is a "mean-teacher" algorithm\cite{miu_GCD}. It uses a "teacher" model to provide pseudo-label supervision and maintains the teacher model through moving averages to reduce the impact of noisy pseudo-labels. GPC is an expectation-maximization-like framework\cite{GPC}. It alternates between representation learning and category count estimation and leverages random splitting and merging mechanisms to dynamically determine prototypes by checking clustering tightness and separability.  DCCL is a dynamic contrastive learning method for GCD tasks\cite{DCCL}. It guides the model to perform contrastive learning on unlabeled data using known category information, enhancing its ability to recognize new categories by dynamically adjusting the selection on positive and negative samples. Spectral open-world representation learning (SORL) provides a graph-theoretical framework for open-world settings\cite{SORL} by using graph factorization to theoretically represent clustering, providing theoretical support and guarantees for practical algorithms.
	
Regularization and active learning methods are also effective in GCD. Spectral open-world representation learning (SORL) provides a graph-theoretical framework for open-world settings\cite{SORL} by using graph factorization to theoretically represent clustering, providing theoretical support and guarantees for practical algorithms. AGCD is an active learning method\cite{AGCD}. It provides an effective way to select a small number of valuable samples for labeling from an "Oracle" to improve the performance of GCD.\par

	\subsection{Vision transformer in GCD}
	Transformer uses multi-head attention mechanisms originally designed for the machine translation task in natural language processing\cite{VIT-1}. In 2020, Cordonnier et al. built upon this by proposing a transformer model for image classification that selects $2\times2$ patches from the input image and applies full self-attention\cite{VIT-2}. In 2021, Alexey Dosovitskiy et al. proposed the vision transformer by applying the transformer architecture directly to image classification\cite{VIT}. Because of its powerful feature representation capability,  
	
	ViT is widely used in GCD as the backbone network or feature-extractor. PromptCAL is a semi-supervised method employing ViT as its backbone for generalized new class discovery (GNCD)\cite{PromptCAL}. It uses contrastive affinity learning in semantic clustering and enhances the semantic discriminative power by embedding learnable visual prompts into the pre-trained ViT and using an auxiliary loss function. SimGCD employs ViT to extract image feature in GCD tasks\cite{simGCD}. It classifies labeled samples with the cross-entropy loss, and distilled them with self-distillation strategies, and employed an entropy regularization term to force the model to predict results with an even entropy distribution across all possible categories. ViT also performs as the backbone in GCD\cite{GCD}, DCCL\cite{DCCL}, GPC\cite{GPC}, PromptCAL\cite{PromptCAL}, and AGCD\cite{AGCD}.
	\par
\section{Preliminaries}\label{sec:preli}
\subsection{Few-shot classification}
FSL primarily aims to train models using limited labeled samples. A typical FSL task is an N-way K-shot classification. N denotes the number of classes. K is the number of labeled samples per class. The K labeled samples constitute a support set, and the rest constitute a query set. A few-shot classification requires the model to classify the query samples based on a very few support samples.

Meta-learning is the most popular FSL strategy currently. It aims at generalizing knowledge across different tasks to tackle new few-shot learning tasks. Meta-learning consists of two stages: meta-training and meta-testing. It meta-trains a model on the base classes with sufficient labeled samples, and meta-tests it on N novel classes with K labeled samples each.

Pre-training and fine-tuning are two popular transfer learning techniques. Their core idea is to use large datasets to train the model to learn general feature representations, transfer these features to the target task, and then fine-tune the model with the limited support samples to adjust the network parameters to fit the target task.

\subsection{Neural field equations}
To describe the effect of changing external inputs on the average activity of the cerebral cortex, we generally use dynamical neural fields as following:
\begin{align}\label{eq:NF}
	\begin{aligned}
		\tau \dot{u}(\mathbf{z},t) &= - u(\mathbf{z},t)+\int_{\Omega}\omega(\mathbf{z} - \mathbf{z}') \phi\big(u(\mathbf{z},t)\big)  \\
		&\quad + s(\mathbf{z},t) .
	\end{aligned}
\end{align}
It is a typical nonlinear integro-differential equation. $ u(z,t)$ denotes the activation at position $z\in\Omega$ and time $t>0$. $\Omega$ is a field in $\mathbb{R}^n$. $s(\mathbf{z},t)$ describe a spatially and temporally variant external input. $\tau>0$ is a time constant.

The integral term $\int_{\Omega}\omega(\mathbf{z} - \mathbf{z}')$ describes the lateral interaction between neurons in the neural field. The interaction kernel $\omega(\cdot)$ determines its strength. Since the lateral interaction is a globally inhibitory and locally excitatory, $\omega(\cdot)$ generally has ``Mexican hat'' shape described by the difference of Gaussian (DoG) functions as follows:
\begin{align}\label{fun:dog}
	\omega_{\sigma}(\mathbf{z}) = a \exp\bigg(-\frac{\left\|\mathbf{z}\right\|^2}{2\sigma^2}  \bigg)
	- b \exp\bigg(-\frac{\left\| \mathbf{z}\right\|^2}{2(3\sigma)^2}\bigg),
\end{align}
where $\left\| \cdot\right\|$ is a vector norm. The constants $a$ and $b$ determines the range of $\omega(\cdot)$. To ensure the maxima of $\omega(\cdot)$ to be $1$, we usually let $a=\dfrac{3}{2},b=\dfrac{1}{2}$. $\sigma$ determines the interaction scale. $\phi(\cdot)$ is a monotonically increasing, non-negative, and bounded activation function given by$$
\phi(u)=\left\{ 
\begin{array}{ll}
	1-\exp({-u}), & u>0 \\
	0, & u\leq0 \\
\end{array}
\right..
$$

Though dynamical neural field theory achieves success in brain science, we may encounter issues in designing a learning method based on it. Firstly, the dynamical neural field equation identifies the input's pattern by the activation induced by it. If two input samples activate a connected region, they belong to the same memory pattern. However, since determining the connectedness of an area in high-dimensional space is difficult, it is impractical to classify feature-extracted image samples. Secondly, since the dynamical neural field equation does not have an analytical solution, we have to solve it with numerical methods, leading to high time and computational cost because of its integration term. Thirdly, for an external input $s(\mathbf{z},t)$ which is positive in a finite region in $\Omega$, the neural field equation may possess a steady state  $u^*_{local}(\mathbf{z})$ with a finite excited region where $u^*_{local}(\mathbf{z})>0$, or an ill-pose steady state $u_{\infty}^*(\mathbf{z})$ called $\infty$-solution that $u_{\infty}^*(\mathbf{z})>0$ for all $\mathbf{z}\in\Omega$, depending on its parameter selection and the input range and strength. Nonetheless, since the condition for generating $\infty$-solution involves integration on the interaction kernel over a region with a variant boundary surface, it is difficult to validate it in high-dimensional space. Finally, since the range of the excited region relies on the scale of lateral interaction, it leads to difficulty in the scale selection since there is little discussion on its selection in high-dimensional space. All these issues make it impractical to build a practical learning model for high-dimensional image data based on the current form of the dynamical neural field equation.

%The dynamical neural field equation (\ref{eq:NF}) is an integro-differential equation.  For an external input $s(\mathbf{z},t)$ which is positive in a finite region in $\Omega$, the neural field equation may possess a steady state  $u^*_{local}(\mathbf{z})$ with a finite excitatory region where $u^*_{local}(\mathbf{z})>0$, or an ill-pose steady state $u_{\infty}^*(\mathbf{z})$ called $\infty$-solution that $u_{\infty}^*(\mathbf{z})>0$ for all $\mathbf{z}\in\Omega$, depending on its parameter selection. Since the discussion on the dynamics of high-dimensional neural field equation is insufficient, the condition to generate the ill-pose steady state $u_{\infty}^*(\mathbf{z})$ is unclear. Moreover, since dynamical neural field equation dose not have analytical solution, we have to solve it with numerical methods, leading to high time and computational cost because of its integration term. Moreover, the dynamical neural field equation store patterns by its excitatory region. In a classification task, if two input samples activate the same connected region, they belongs to the same class. However, since determining the range of connected regions in high-dimensional space is quite difficult, it is impractical to classify feature extracted image vectors in this way.

%For these issue, we 

\section{Method}\label{sec:method}
%In this section, we will introduce a new neural field-based architecture for classification and provide some learning strategies to expand its FSL applicability and computational efficiency.
\subsection{Feature extraction and preprocessing}
In a typical ViT, the linear projection, patch+position embedding, and transformer encoder constitute its feature extractor, as shown in Figure \ref{fig:structure}a. To simplify the discussion, we denote the effect of the feature extractor by the following equation: 
\begin{equation*}
	\mathbf{x}=ViT_{fe}(\mathbf{Z}).	
\end{equation*}
where $\mathbf{Z}$ is an image and $\mathbf{x}$ is its feature vector. In this way, the function of the feature extractor is a map from an image space $\mathbf{I}$ to a feature space $\mathbf{\Omega}$. 

The extracted feature $\mathbf{x}$ is a high-dimensional vector. It usually contains some redundant information useless for classification. The redundant information may have negative impact on the classification accuracy and cost more computation resources. Therefore, we employ dimensional reduction methods to reduce the feature dimensions. We describe these processes by the following function: 
\begin{equation*}
	\mathbf{z}=R_d(\mathbf{x}),	
\end{equation*}
where $\mathbf{z}$ is the dimensional reduced feature vector.

\subsection{Static neural field function}
When we identify a query sample's class based on a dynamical neural field, we shall check whether it can generate a connected excited region with some support samples. Since it is difficult to validate, we propose a soft condition: whether the query sample can activate the neurons corresponding to some support samples. If it can activate these neurons, it can also generate a connected excited region with the corresponding support samples. In this way, we change the requirement from determining the connectedness of an excited region to checking the activation of several specific neurons. 

Nevertheless, we still have to carefully choose the parameters of the dynamical neural field equation to avoid $\infty$-solution and solve it numerically. For this issue, observing that the solution with a finite excitatory region can be approximate by the convolution on the input function $s(\mathbf{z},t)$ interaction kernel $\omega_{\sigma}(\cdot)$ with a proper scale when the input is static $s(\mathbf{z},t)=s(\mathbf{z})$, we propose a function to describe the activation of neural field as follows:
\begin{equation}\label{fun:snf}
	u(\mathbf{z})=\phi\bigg(\int_{\Omega}\omega_{\sigma}(\mathbf{z} - \mathbf{z}') \phi\big(s(\mathbf{z})\big)\bigg).
\end{equation}	
This function can generate a similar shape of the excited region with a dynamical neural field and will not generate the ill-posed $\infty$-solution.

The positive activation in a dynamical neural field will impact its subsequent evolution. We need to calculate all the neurons in the field since they synchronously receive lateral activation from all their neighbors. However, in the proposed static neural field, the activation is determined by its distance to the input. Therefore, we can compute the activation of some specific neurons and simplify $u(\mathbf{z})$ into a discrete one: 
\begin{equation}\label{fun:dnf}
	u_k=\phi\big(\omega_{\sigma}(\mathbf{z}_k - \mathbf{z}_s) \phi(s)\big), k=1,2,\cdots,
\end{equation}
where $u_k$ is the activation of a neuron and $\mathbf{z}_k$ is its position in neural field. $s$ is the external input corresponding to the query sample and $\mathbf{z}_s$ is its position.  It significantly reduces the computation and parameter selection complexity in leveraging the neural field. The remaining issue is how to store and identify sample information.

\subsection{Architecture of NF classifier}
Suppose $\mathcal{S}=\{\mathbf{Z}_1,\mathbf{Z}_2,\cdots,\mathbf{Z}_m\}$ is the support set and $\mathcal{L}=\{l_1,l_2,\cdots,l_m\}$ is the set of labels whose values are $\{y_1,y_2,\cdots,y_{m_c}\}$. We extract their features by $$\mathbf{z}_i=ViT_{fe}(\mathbf{Z}_i),$$ and then obtain the feature vectors of the support sample $\{\mathbf{z}_1,\mathbf{z}_2,\cdots,\mathbf{z}_m\}$. To store the information of these feature vectors, we use a static neural field to memorize their positions in feature space. We call it an elementary field and describe the neuronal activation corresponding to the support samples by the following equations: 
\begin{equation}\label{fun:dnf-s}
	u_i=\phi\big(\omega_{\sigma}(\mathbf{z}_i - \mathbf{z}_q) \phi(s_q)\big), i=1,2,\cdots,m,
\end{equation}
where $u_i$ is the activation of the neuron corresponding to the $i$th support sample and $\mathbf{z}_i$ is its position in neural field. $s_q=1$ is the external input corresponding to the query sample and $\mathbf{z}_q$ is its position. When we input a query sample into this field, these neurons will receive its excitatory or inhibitory impact determined by their distance and the interaction scale. 

To memorize the class information of the support samples, we design another field with neurons corresponding to the class labels and refer it to the high-level field. The high-level field contains $s$ neurons. Each one corresponds to a class label $y_j$, $j=1,2,\cdots,m_c$. We describe their response to the input from the elementary field by the following equations: 
\begin{equation}
	\begin{aligned}
		v_{j}&=\phi\big(\sum_{i=1}^{m}w_{j,i}u_{i}
		\big)
		\\&=\phi\bigg(\sum_{i=1}^{m}w_{j,i}\phi\big(\omega_{\sigma}(\mathbf{z}_i-\mathbf{z}_k)\phi(s_q)\big)\bigg).
	\end{aligned}			
\end{equation}
Suppose the neurons corresponding to different data classes are distant. We ignore the lateral interaction between them to reduce computational cost since it has almost no impact on the classification result. $w_{j,i}$ is the weight of cross-field connection from the $i$th elementary neuron to the $j$th high-level one. If the $i$th support sample's label is $y_j$, we let $w_{j,i}=1$, else, $w_{j,i}=0$. In this way, if an input sample activates any elementary neurons with the label $y_j$, the cross-field connection will transfer their positive activation and activate the $j$th high-level neuron. Therefore, we can classify the input sample by checking the activation of the high-level field. If the activation of the $j$th high-level neuron is positive in prediction, we will label the input sample by $y_j$.

The NF classification has a specific advantage in GCD. When we detect a new category, we add an elementary neuron corresponding to the input sample and a high-level neuron corresponding to its category and connect them with a cross-field connection. Since this operation has no impact on the other neurons and their connections, it will never lead to catastrophic forgetting. Therefore, we replace the MLP head with the neural field-based classifier and obtain the modified architecture ViTNF, as shown in Figure \ref{fig:structure}b.

\subsection{Parameter selection}
The lateral interaction scale $\sigma$ plays a critical role in the prediction. When it is too small, the range of its excitatory lateral interaction of the elementary field is insufficient to activate any elementary neurons, so we cannot find any activated neurons in the high-level field. When it is too large, the excitatory range may cover the elementary neurons connected to different high-level neurons, and then we will find more than one activated neuron. We cannot determine the category of the input sample in both cases and need to find a way to deal with these situations. 

Observing that the excitatory range is monotonously increasing with $\sigma$, for general few-shot classification, we can adjust $\sigma$ with a simple strategy: when there is not any activated high-level neuron, we increase it; when there is more than one, we decrease it. Whenever we find a small $\sigma$ inducing the former case and another one leading to the latter case denoted by $\sigma_{min}$ and $\sigma_{max}$, we can find the proper scale between them. However, when there are unknown categories, if we continuously increase the $\sigma$, the input sample will activate a high-level neuron corresponding to a known category. Therefore, we cannot detect any new category in this way.

An effective way to solve this issue is to find a proper interaction scale $\sigma$ that describes the sample distribution in the same category. Whenever an input sample cannot activate any high-level neuron, we classify it into a new category. Samples in different datasets have their specific distribution scales. Therefore, we standardize the sample feature vectors in the same dataset and still denote them by $\mathbf{z}$. To simply the discussion, we propose the following assumptions:  
\begin{assumption}\label{ap:1}
	The samples in the same category follow a symmetric distribution in the feature space $\Omega$.	
\end{assumption}

\begin{assumption}\label{ap:2}
	The samples in the different categories share a similar distribution scale in the feature space $\Omega$.	
\end{assumption}

\begin{assumption}\label{ap:3}
	The samples in the different categories are separable in the feature space $\Omega$.	
\end{assumption}

These assumptions are general in statistical analysis. Though the original image samples may not satisfy them, their feature vectors can meet these assumptions in most cases, attributed to the ViT's powerful representation capability. Indeed, we propose them for convenience in discussion, and they are not strict restrictions in practical applications.  

It is easy to prove the following result:
\begin{proposition}\label{pr:radius}
	Suppose the two separable hyperspheres share the same radius $r_s$ in a large hypersphere $B$ with radius $r$, then $r_s\leq r/2$. 
\end{proposition}
For a standardized sample set, the sample variance is $1$. Ignoring some extreme outliers, most of its samples are within the hypersphere $B$ with radius $r=3$. Following Assumptions \ref{ap:1}, the samples in the $j$th category are also distributed in a hypersphere $B_j\subset B$. According to Assumptions \ref{ap:2} to \ref{ap:3} and Proposition \ref{pr:radius}, the radius of $B_j, j=1,2,\cdots,s$ is no more than $1.5$.

The interaction kernel determines the range of the excitatory region induced by an input sample. To analyze its property, we have the following result:
\begin{proposition}
	When the interaction kernel $\omega_\sigma(\cdot)$ is a DoG function defined by (\ref{fun:dog}), its excitatory radius $r_e=\dfrac{3\sqrt{\ln(A/B)}}{2}$.
\end{proposition}
We can prove it by solving the following equation:
\begin{align}
	\omega_{\sigma}(\mathbf{z}) =0.
\end{align}
When $a=\dfrac{3}{2}$ and $b=\dfrac{1}{2}$, the excitatory radius $r_e=\|\mathbf{z}\|=\dfrac{3\sqrt{\ln(3)}}{2}\sigma$. if $\sigma=1$, we can obtain $r_e=\dfrac{3\sqrt{\ln(3)}}{2}=1.57$, just slightly larger than $1.5$. Therefore, it is proper to let the upper bound of $\sigma$ be $1$.

The number of reserved dimensions indicates the representation capacity of the reduced feature space. We present a way to evaluate its capacity:
\begin{proposition}\label{pr:cap}
	A $n$-dimensional hypersphere $B$ with radius $3$ can contain $2n+1$ separable unit open hyperspheres in it at least. 
\end{proposition}
\begin{proof}
	Suppose the center of $B$ is the origin $O$. Let $O_1^{\pm}= (\pm2,0,\cdots,0),O_2^{\pm}=(0,\pm2,\cdots,0),\cdots,O_n^{\pm}=(0,0,\cdots,\pm2)$ and the origin $O$ be the centers of $2n+1$ unit hyperspheres. Since the distance is $2\sqrt{2}$ between the points on different axes and $4$ on the same axis, the hyperspheres centered at $O_k^{\pm}, k=1,2,\cdots,n$ are separated. It is easy to see that they do not intersect the unit hypersphere at $O$. Therefore, $B$ can contain $2n+1$  separable unit open hyperspheres in it. 
\end{proof}
Though the actual sample distribution may not satisfy Assumptions \ref{ap:1} to \ref{ap:3} in a practical application, Proposition \ref{pr:cap} still provides an applicable criterion for the feature reduction.

\subsection{Algorithms for GCD}
\begin{figure*}[htb]
	\centering
		\includegraphics[width=0.7\textwidth]{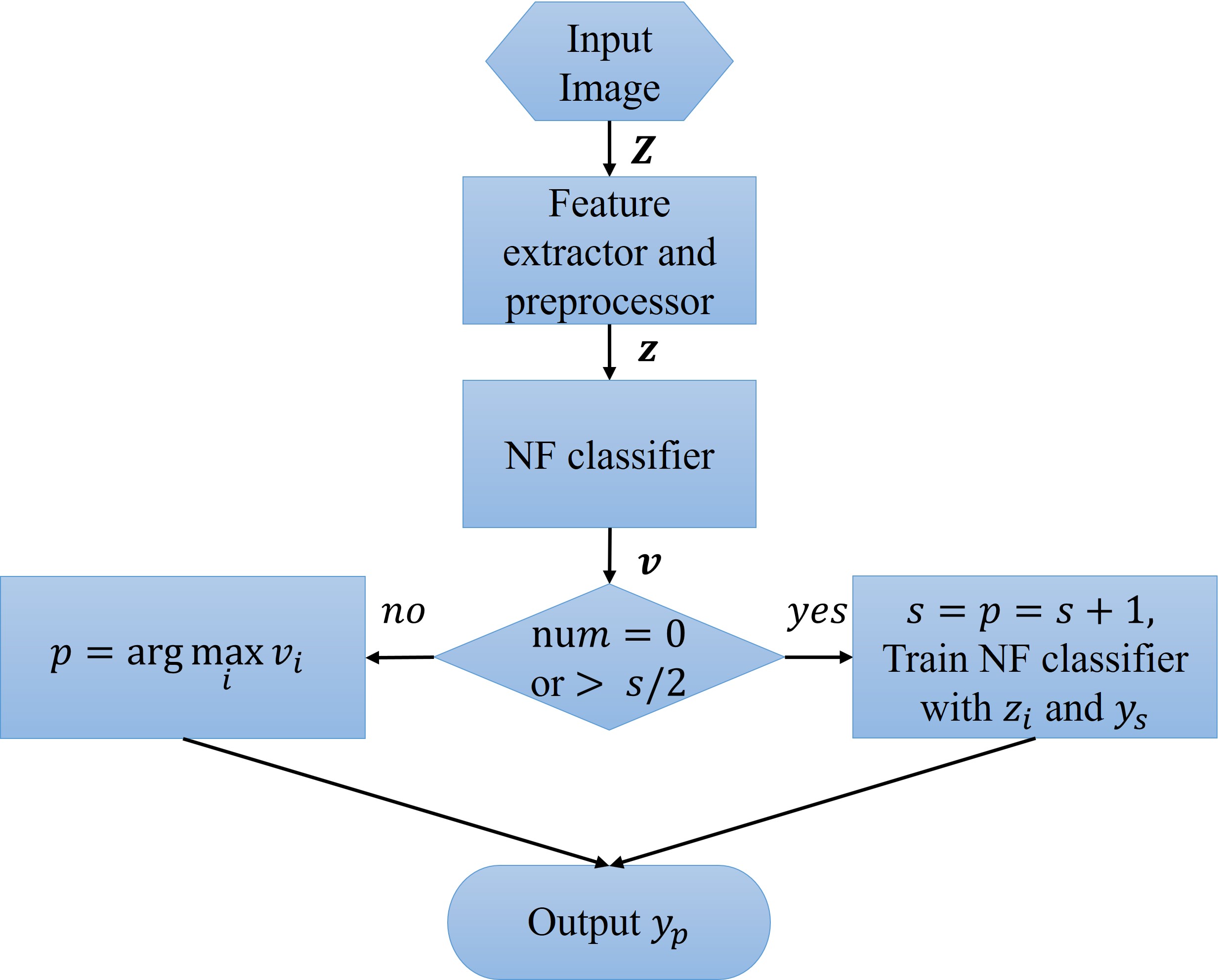}
	\caption{The prediction process of ViTNF. }\label{fig:predict}
\end{figure*}

We can classify an input sample according to the high-level neuron activation vector $\mathbf{v}=\{v_1,v_2,\cdots,v_s\}$. If an input sample cannot activate any high-level neuron when $\sigma=1$, we classify it to a new category, provide it a Pseudo-label, and train the NF classifier with it. If it activates multiple high-level neurons, we adapt $\sigma$ following Algorithm \ref{alg1}, where $s$ is the total number of old and detected categories, $0<\lambda<1$ an iteration ratio constant, $num$ the number of positive high-level neurons, and $p$ the sequence number of the predicted category. Since the length of the interval $(\sigma_{min},\sigma_{max})$ is less than $100\lambda$ percentages of the previous step, the range of adjustment at the $k$ step is no more than $\lambda^k$, inducing the change in excitatory radius less than $1.57\lambda^k$. When $k$ is large, the change becomes too small to continue iterating. Therefore, we set a terminal number $T$ to stop the iteration. When it still has activated high-level neurons at the terminal, we assign the input sample the category of the high-level neuron with the highest activation when $num\leq s/2$ or a new one $y_{s+1}$ when $num> s/2 $. When detecting new category $y_{s+1}$, we let $s=s+1$, train the NF classifier by adding an elementary neuron corresponding to $\mathbf{z}$ and a high-level neuron corresponding to $y_s$. The whole prediction process is shown in Figure \ref{fig:predict}.
\begin{algorithm}
	\caption{Prediction Algorithm for NF Classifier}\label{alg1}
	\begin{algorithmic}[1]
		\State \textbf{Input:} $\mathbf{z},s,T$
		\State \textbf{Output:} $p$
		\State Initialize $\sigma_{min}=0$, $\sigma_{max}=1$, $\sigma=1$, $\lambda$, $num=0$;
		\State Calculate $\mathbf{v}$ and update $num$;
		\If{$num =0 $}
		\State $p=s+1$;
		\Else
		\For{k=1:T}
		\If{$num > 1$}
		\State $\sigma_{max}=\sigma$, $\sigma=\sigma_{min}+\lambda(\sigma_{max}-\sigma_{min})$;
		\EndIf
		\If{$num =0$}
		\State $\sigma_{min}=\sigma$, $\sigma=\sigma_{max}-\lambda(\sigma_{max}-\sigma_{min})$;
		\EndIf
		\State Calculate $\mathbf{v}$ and update $num$;		
		\If{$num = 1$}
		\State Break;
		\EndIf
		\EndFor
		\If{$1\leq num\leq s/2$} 
		\State $p=\arg\max_{j=1}^{s}v_j$;
		\EndIf
		\If{$num>s/2$}
		\State $p=s+1$; 
		\EndIf
		\EndIf
	\end{algorithmic}
\end{algorithm}

	\section{Experiment}\label{Experiment}
	In this section, we compare the state-of-the-art GCD models with our proposed ViTNF across five different datasets to demonstrate its superiority. Additionally, through ablation experiments, we analyze the impact of parameter and distance metrics selection.
	
	\subsection{Datasets and experimental setting}
	To validate the effectiveness of ViTNF in GCD, we test it on CIFAR-10\cite{Cifar10}, CIFAR-100\cite{Cifar10}, ImageNet-100\cite{Imagenet100}, CUB200-2011\cite{CUB_200_2011}, and Stanford Cars\cite{SC} as the datasets for our comparative study. We show the details of these datasets as follows and summarize them in Table \ref{dataset}.
	
	\begin{itemize}
		\item[$\bullet$] CIFAR-10: This dataset consists of 60,000 color images of size 32×32, divided into 10 categories, 50,000 images for training and 10,000 images for testing.
		\item[$\bullet$]CIFAR-100: This dataset contains images from 100 categories, with each category having 600 images. Among them, 500 images per category are used for training, while the remaining 100 images form the test set. Each image has two labels: a superclass label and a subclass label, corresponding to its broad category and specific category, respectively.
		\item[$\bullet$]ImageNet-100: This dataset is a subset of ImageNet with 100 categories. The training set contains 126,689 images, while the test set includes 5,000 images.
		\item[$\bullet$] CUB200-2011: This dataset contains a total of 11,788 bird images divided into 200 categories. The training set includes 5,994 images, while the test set consists of 5,794 images.
		\item[$\bullet$] Stanford Cars: This dataset contains a total of 16,185 images of different car models, divided into 196 categories, 8,144 images for training and 8,041 images for testing.
	\end{itemize}
	
\begin{table}[h]
	\caption{The details of CIFAR-10, CIFAR-100, ImageNet-100, CUB200-2011, and Stanford Cars.}
	\label{dataset}
	\centering
	\footnotesize
	\begin{tabularx}{\textwidth}{c|XXXXX}
		\toprule
		& \textbf{CIFAR-10} & \textbf{CIFAR-100} & \textbf{ImageNet-100} & \textbf{CUB200-2011} & \textbf{Stanford Cars} \\
		\midrule
		Total classes & 10 & 100 & 100 & 200 & 196 \\
		Total images & 60,000 & 60,000 & 131,689 & 11,788 & 16,185 \\
		Training sets & 50,000 & 50,000 & 126,689 & 5,994 & 8,144 \\
		Validation sets & 10,000 & 10,000 & 5,000 & 5,794 & 8,041 \\
		\bottomrule
	\end{tabularx}
\end{table}

	We use ViT-B/16 as a backbone pre-trained for 10 epochs on the I21K dataset\cite{VIT}. The iterative terminal number $T=4$ and the ratio $\lambda=0.4$ for all tests. 
	
	We use a typical sample selection setting for GCD as follows: 
\begin{itemize}
	\item[$\bullet$] Old classes: following 5-way 10-shot setting, that is, randomly select five categories for meta-testing, 10 samples per class to form the support set, and group the rest samples to the test set.
	\item[$\bullet$] New classes: randomly select another five categories besides the old ones.	
\end{itemize}
Since we do not meta-train ViTNF on the target dataset, we randomly select the old and new classes from all its classes.

To preserve the local structure of sample distribution, we use Laplacian eigenmaps (LE) to reduce the sample dimensions. Since the number of sample categories is 10, according to Proposition \ref{pr:cap}, we reserve the four dimensions with the lowest positive eigenvalues. We use the Euclidean distance to measure the dissimilarity of the reduced feature vectors and present the average results over 600 epochs. 
	
	\subsection{Comparison with state-of-the-art models' results}
We compare ViTNF with openLDN\cite{openLDN}, ORCA\cite{orca}, simGCD\cite{simGCD}, GCD\cite{GCD}, DCCL\cite{DCCL}, GPC\cite{GPC}, PromptCAL\cite{PromptCAL}, SORL\cite{SORL}, and AGCD\cite{AGCD}. In this experiment, we use three different accuracies: accuracy in all classes (referred to as ``All''), accuracy in the new classes (referred to as ``New''), and accuracy in the old class detection (referred to as ``Old''). We present the results obtained on CIFAR-10, CIFAR-100, and ImageNet-100 in Table \ref{result1}, obtained on the CUB-200 and Stanford Cars datasets in Table \ref{result2}, with the best result for each task highlighted in bold, and the second-best result underlined.

The experimental results in Table \ref{result2} demonstrate that ViTNF achieves state-of-the-art performance across all three benchmark datasets (CIFAR-10, CIFAR-100, ImageNet-100), significantly outperforming existing methods in both old and new class accuracy, as well as all class accuracy (All). Notably, ViTNF exhibits remarkable consistency and stability, as evidenced by its minimal standard deviations (±0.0003–0.0023), which are orders of magnitude smaller than those of other methods.

On CIFAR-10, ViTNF achieves 99.0\% Old, 97.5\% New, and 98.3\% All accuracy, surpassing the best-performing baselines in old and all classes (e.g., PromptCAL: 96.6\% Old, 97.9\% All), slightly lower than simGCD (98.1\%) and PromptCAL (98.5\%) by -0.6\% and -1\% in new classes. 

On CIFAR-100, ViTNF achieves dramatic accuracy: 99.3\% (Old), 97.8\% (New), and 98.6\% (All), surpassing the best competitor (GPC) by +14.7\% in old classes and outperforming simGCD by +20.0\% in new ones.

For ImageNet-100, ViTNF achieves dramatic accuracy (99.8\% Old, 98.6\% New, 99.2\% All), exceeding AGCD (the second-best in All) by +9.6\% Old, +22.1\% New, and +15.9\% All, highlighting its scalability to complex datasets.

On CUB-200, ViTNF achieves unprecedented scores: 95.3\% (Old), 92.3\% (New), and 94.1\% (All), surpassing the strongest baseline ($\mu$GCD: 74.0\% All) by +20.1\%. This leap underscores its ability to resolve subtle inter-class variations in fine-grained tasks.

For Stanford Cars, ViTNF delivers 90.5\% (Old), 90.1\% (New), and 90.3\% (All) accuracy, outperforming $\mu$GCD (76.1\% All) by +14.2\% in all class accuracy, and just slightly lower than $\mu$GCD (91.01\%) by -0.5\% in old class accuracy. Notably, the new class accuracy (90.1\%) of ViTNF exceeds $\mu$GCD (68.9\%) by +21.2\%, reflecting its exceptional novelty discovery capability.

Unlike methods such as GCD or GPC achieving a significant decline between old and new class accuracy (e.g., GCD: 89.8\% Old vs. 66.3\% New on ImageNet-100), ViTNF maintains harmoniously high accuracy across both categories on all datasets, indicating its robustness in generalized category discovery without overfitting to known or novel classes.

The negligible standard deviations (e.g., ±0.0003 for old classes in ImageNet-100) underscore ViTNF’s reliability, exhibiting higher variance in results.

While all ViT-based methods (GCD, simGCD, etc.) share similar transformer encoders, ViTNF more fully leverages the powerful feature extraction capabilities of ViT, enabling it to outperform even strong baselines like PromptCAL (+1.4\% all class accuracy on CIFAR-10) and $\mu$GCD (+19.2\% New accuracy on CIFAR-100).
	
%	  On the dataset CIFAR-10, we can see that the proposed ViTNF achieves the highest accuracy of $0.990$ in old class detection and $0.983$ in the new-class detection, and the third best accuracy $0.975$ in new class detection, $0.006$ and $0.01$ lower than simGCD ($0.981$) and PromptCAL ($0.985$). 
	 
%	 On CIFAR-100, the proposed ViTNF achieves significantly higher accuracy than all the other methods. It achieves an accuracy of $0.993$ in the old-class recognition task with $15$ percentages' leading to the second best GPC ($0.846$), an accuracy of $0.978$ in the new class detection with $20$ percentages' leading to second-best simGCD($0.778$), and an accuracy $0.986$ in the overall detection with $27$ percentages' leading to the second-ranked PromptCAL($0.812$).\par

%	 On ImageNet-100, the proposed ViTNF also achieves significantly higher accuracy than all the other methods.It achieves an accuracy of $0.998$  in the old-class recognition task with $6$ percentages' leading to the second best GPC ($0.934$), an accuracy of $0.986$  in the new class detection with $20$ percentages' leading to second-best PromptCAL ($0.783$),  and an accuracy $0.986$ in the overall detection with $27$ percentages' leading to the second-ranked AGCD achieves $0.833$. 

	\begin{table}[h]
		\caption{The accuarcy of ViTNF on CIFAR-10, CIFAR100, ImageNet-100.}
		\label{result1}
		\centering
		\resizebox{\linewidth}{!}{%
			\begin{tabular}{c|ccc|ccc|ccc}
				\toprule
				\multirow{2}{*}{\textbf{Model(Encoder)}} & \multicolumn{3}{c|}{\textbf{CIFAR-10}} & \multicolumn{3}{c|}{\textbf{CIFAR-100}} & \multicolumn{3}{c}{\textbf{ImageNet-100}} \\
				\cline{2-10}
				& Old& New & All& Old& New & All & Old& New & All \\
				\midrule  
				openLDN(Resnet18) & 0.957 & 0.951 & 0.954 & 0.741 & 0.445 & 0.593 &0.896&0.686&0.791 \\
				ORCA(Resnet18)    &0.882  &0.904  &0.897  &0.669  &0.430  &0.481  &0.891  &0.721  &0.778  \\
				SORL(Resnet18)    &0.940  &0.925  &0.935  &0.682  &0.520  &0.561  &\textbackslash&\textbackslash&\textbackslash\\
				GCD(VIT)     &\underline{0.979}	&0.882  &0.915  &0.762  &0.665  &0.730  &0.898  &0.663  &0.741\\
				simGCD(VIT)  &0.951  &\underline{0.981}  &0.971  &0.812  &\underline{0.778}  &0.801  &0.931  &0.779  &0.830\\
				DCCL(VIT)    &0.965  &0.969  &0.963  &0.768  &0.702  &0.753  &0.905  &0.762  &0.805\\
				GPC(VIT)     &0.976  &0.870  &0.906  &\underline{0.846}  &0.601  &0.754  &\underline{0.934}  &0.667  &0.753\\
				PromptCAL(VIT) &0.966&\textbf{0.985}&\underline{0.979}  &0.842  &0.753  &\underline{0.812}  &0.927  &\underline{0.783}  &0.831\\
				AGCD(VIT)    &0.946  &0.928  &0.932  &0.757  &0.668  &0.713  &0.902  &0.765  &\underline{0.833}\\
				\midrule
				ViTNF(VIT)    &$\textbf{0.990}\pm0.0007$  &$0.975\pm0.0023$  &$\textbf{0.983}\pm0.0011$  &$\textbf{0.993}\pm0.0006$  &$\textbf{0.978}\pm0.0019$  &$\textbf{0.986}\pm0.0009$  &$\textbf{0.998}\pm0.0003$  &$\textbf{0.986}\pm0.0013$  &$\textbf{0.992}\pm0.0006$\\
				\bottomrule
			\end{tabular}%
		}
	\end{table}

	\begin{table}[h]
		\caption{The accuarcy of ViTNF on CUB-200 and Stanford Cars.}
		\label{result2}
		\centering
		\resizebox{\linewidth}{!}{%
			\begin{tabular}{c|ccc|ccc}
				\toprule
				\multirow{2}{*}{\textbf{Model(Encoder)}} & \multicolumn{3}{c|}{\textbf{CUB-200}} & \multicolumn{3}{c}{\textbf{Stanford Cars}} \\
				\cline{2-7}
				& Old& New & All& Old& New & All\\
				\midrule  
				GCD(VIT)     &0.566	&0.487  &0.513  &0.576  &0.299  &0.390  \\
				simGCD(VIT)  &0.656  &0.577 &0.603  &0.719  &0.450  &0.538 \\
				DCCL(VIT)    &0.608  &0.649  &0.635  &0.557  &0.362  &0.431  \\
				GPC(VIT)     &0.555  &0.475  &0.520  &0.589  &0.274  &0.382 \\
				PromptCAL(VIT) &0.644&0.621  &0.629  &0.701  &0.406  &0.502  \\
				AGCD(VIT)    &0.665  &0.667  &0.666  &0.577  &0.393  &0.484  \\
				$\mu$GCD(VIT)&\underline{0.759}  &\underline{0.731}  &\underline{0.740}  &\textbf{0.910}  &\underline{0.689}  &\underline{0.761}\\
				\midrule
				ViTNF(VIT)    &$\textbf{0.953}\pm0.0030$ &$\textbf{0.923}\pm0.0048$ &$\textbf{0.941}\pm0.0027$ &$\underline{0.905}\pm0.0033$ &$\textbf{0.901}\pm0.0038$ &$\textbf{0.903}\pm0.0024$\\
				\bottomrule
			\end{tabular}%
		}
	\end{table}
	The experimental results demonstrate the dramatic performance of ViTNF in generalized category discovery. It combines a pre-trained ViT feature extractor with the proposed NF classifier, introducing significant advantages in training efficiency and state-of-the-art accuracy while maintaining exceptional stability without meta-training or fine-tuning.

	\subsection{Ablation studies}
The criterion for detecting a new category in Algorithm \ref{alg1} is the number $num$ of activated neurons. To analyze its impact, we choose different values and check the obtained accuracy in Old, New, and All classes with Euclidean distance (Euc), cosine distance (Cos), and Mahalanobis distance (Mah). 

The ablation studies reveal critical insights into the performance of ViTNF under varying configurations of the parameter $num$ (proportional to sample size) and distance metrics (Euclidean, Cosine, Mahalanobis). The results demonstrate that ViTNF achieves optimal performance when using $num = \dfrac{s}{2}$ with the Euclidean (Euc) metric, establishing it as the most robust and effective configuration across all datasets.

With $num = \dfrac{s}{2}$ and Euc, ViTNF attains peak performance: 99.0\% (Old), 97.5\% (New), and 98.3\% (All) on CIFAR-10, 99.3\% (Old), 97.8\% (New), and 98.6\% (All) on CIFAR-100, and 99.8\% (Old), 98.6\% (New), and 99.2\% (All) on ImageNet-100,
outperforming other values (e.g., $\dfrac{3s}{4}$ or $\dfrac{2s}{3}$)  in all class accuracy with highlighting its ability to balance old and new class discrimination. Similarly, $num = \dfrac{s}{2}$ with Euc achieves 94.1\% (All) on CUB-200 and 90.3\% (All) on Stanford Cars, surpassing other configurations by +1.8–28.6\%. Notably, ViTNF’s new class accuracy on Stanford Cars (90.1\%) nearly matches its Old class performance (90.5\%), eliminating bias toward known categories. The $num = \dfrac{s}{2}$ configuration maintains high performance across both coarse-grained (CIFAR, ImageNet) and fine-grained (CUB-200, Stanford Cars) datasets, proving its scalability.

Distance metrics also have a significant impact on the accuracy. Euclidean distance consistently delivers the highest accuracy and stability (e.g., ±0.0003–0.0033 deviations). Its success suggests that geometric feature separation is optimal for ViTNF. Despite excellent old class accuracy (e.g., 95.9\% on CUB-200), Cosine distance catastrophically fails on new classes (33.6–53.2\%), causing drastic drops in all class accuracy(e.g., 71.9\% on CUB-200 vs. 94.1\% with Euc), highlighting the unsuitability of angular similarity for novelty discovery in ViTNF. Mah performs moderately and lags behind Euc by 1.0–3.0\% in all class accuracy.

	\par
	\begin{table}[h]
		\caption{Results on CIFAR-10, CIFAR-100, and ImageNet-100 with various $num$ and distance metrics.}
		\label{aresult1}
		\centering
		\resizebox{\linewidth}{!}{%
			\begin{tabular}{c|c|ccc|ccc|ccc}
				\toprule
				\multirow{2}{*}{\textbf{$num$}} & \multirow{2}{*}{\textbf{Metric}} & \multicolumn{3}{c|}{\textbf{CIFAR-10}} & \multicolumn{3}{c|}{\textbf{CIFAR-100}} & \multicolumn{3}{c}{\textbf{ImageNet-100}} \\
				\cline{3-11}
				& & Old & New & All & Old & New & All & Old & New & All \\
				\midrule  
				\multirow{3}{*}{\centering \textbf{$\dfrac{3s}{4}$}}
				& Euc & $0.984 \pm 0.0008$ & $\underline{0.938} \pm 0.0021$ & $\underline{0.962} \pm 0.0011$ & $0.987\pm 0.0008$ & $\underline{0.953} \pm 0.0022$ & $\underline{0.971} \pm 0.0011$ & $0.987 \pm 0.0004$ & $\underline{0.952} \pm 0.0019$ & $\underline{0.971} \pm 0.0009$ \\
				& Cos & $0.983 \pm 0.0098$ & $0.362 \pm 0.0860$ & $0.683 \pm 0.0411$ & $0.984 \pm 0.0084$ & $0.329 \pm 0.0800$ & $0.667 \pm 0.0432$ & $0.986 \pm 0.0033$ & $0.350 \pm 0.0610$ & $0.679 \pm 0.0270$ \\
				& Mah & $0.985 \pm 0.0007$ & $0.930 \pm 0.0024$ & $0.960 \pm 0.0012$ & $0.987\pm 0.0007$ & $0.951 \pm 0.0022$ & $0.969 \pm 0.0011$ & $0.988 \pm 0.0004$ & $0.947 \pm 0.0017$ & $0.969 \pm 0.0008$ \\
				\midrule  
				\multirow{3}{*}{\centering \textbf{$\dfrac{2s}{3}$}}
				& Euc & $0.984 \pm 0.0008$ & $0.936\pm 0.0023$ & $\underline{0.962} \pm 0.0011$ & $0.987\pm 0.0007$ & $\underline{0.953} \pm 0.0021$ & $\underline{0.971} \pm 0.0011$ & $0.988 \pm 0.0003$ & $0.951\pm 0.0019$ & $0.970 \pm 0.0009$ \\
				& Cos & $0.983 \pm 0.0084$ & $0.456 \pm 0.0788$ & $0.732 \pm 0.0432$ & $0.986 \pm 0.0074$ & $0.429 \pm 0.0786$ & $0.723 \pm 0.0404$ & $0.988 \pm 0.0051$ & $0.420 \pm 0.0559$ & $0.721 \pm 0.0293$ \\
				& Mah & $\underline{0.986} \pm 0.0007$ & $0.934 \pm 0.0024$ & $\underline{0.962} \pm 0.0012$ & $0.987\pm 0.0007$ & $0.952 \pm 0.0020$ & $0.970 \pm 0.0010$ & $0.988\pm 0.0004$ & $0.951\pm 0.0015$ & $\underline{0.971} \pm 0.0008$ \\
				\midrule  
				\multirow{3}{*}{\centering \textbf{$\dfrac{s}{2}$}}
				& Euc & $\textbf{0.990}\pm0.0007$  &$\textbf{0.975}\pm0.0023$  &$\textbf{0.983}\pm0.0011$  &$\textbf{0.993}\pm0.0006$  &$\textbf{0.978}\pm0.0019$  &$\textbf{0.986}\pm0.0009$  &$\textbf{0.998}\pm0.0003$  &$\textbf{0.986}\pm0.0013$  &$\textbf{0.992}\pm0.0006$ \\
				& Cos & $\underline{0.986} \pm 0.0098$ & $0.532 \pm 0.0790$ & $0.773 \pm 0.0411$ & $\underline{0.988} \pm 0.0077$ & $0.515 \pm 0.0839$ & $0.768 \pm 0.0660$ & $\underline{0.989} \pm 0.0064$ & $0.498 \pm 0.0745$ & $0.760 \pm 0.0320$ \\
				& Mah & $0.982 \pm 0.0008$ & $\underline{0.938} \pm 0.0022$ & $\underline{0.962} \pm 0.0011$ & $0.986 \pm 0.0007$ & $0.952 \pm 0.0022$ & $0.970 \pm 0.0011$ & $0.988\pm 0.0004$ & $0.949 \pm 0.0015$ & $0.970\pm 0.0008$ \\
				\bottomrule
			\end{tabular}%
		}
	\end{table}

	\begin{table}[h]
		\caption{Results on CUB-200 and Stanford Cars with various $num$ and distance metrics.}
		\label{aresult2}
		\centering
		\resizebox{\linewidth}{!}{%
			\begin{tabular}{c|c|ccc|ccc}
				\toprule
				\multirow{2}{*}{\textbf{$num$}} & \multirow{2}{*}{\textbf{Metric}} & \multicolumn{3}{c|}{\textbf{CUB-200}} & \multicolumn{3}{c}{\textbf{Stanford Cars}} \\
				\cline{3-8}
				& & Old & New & All & Old & New & All \\
				\midrule  
				\multirow{3}{*}{$\dfrac{3s}{4}$}
				& Euc & $0.935 \pm 0.0033$ & $\underline{0.908} \pm 0.0048$ & $\underline{0.923} \pm 0.0027$ & $0.896 \pm 0.0034$ & $0.820 \pm 0.0042$ & $0.863\pm 0.0026$ \\
				& Cos & $\underline{0.957} \pm 0.0211$ & $0.336 \pm 0.0621$ & $0.655 \pm 0.0281$ & $\textbf{0.917} \pm 0.0403$ & $0.428 \pm 0.0532$ & $0.691 \pm 0.0503$ \\
				& Mah & $0.940 \pm 0.0035$ & $0.877 \pm 0.0046$ & $0.910 \pm 0.0026$ & $0.893 \pm 0.0032$ & $0.814 \pm 0.0040$ & $0.859 \pm 0.0024$ \\
				\midrule  
				\multirow{3}{*}{$\dfrac{2s}{3}$}
				& Euc & $0.936 \pm 0.0032$ & $0.907 \pm 0.0047$ & $\underline{0.923} \pm 0.0027$ & $0.898 \pm 0.0034$ & $0.819 \pm 0.0043$ & $0.863\pm 0.0025$ \\
				& Cos & $0.956 \pm 0.0157$ & $0.376 \pm 0.0998$ & $0.676 \pm 0.0449$ & $\underline{0.916} \pm 0.0486$ & $0.439 \pm 0.0570$ & $0.698 \pm 0.0350$ \\
				& Mah & $0.940 \pm 0.0032$ & $0.874 \pm 0.0053$ & $0.909 \pm 0.0029$ & $0.898 \pm 0.0033$ & $0.811 \pm 0.0042$ & $0.860 \pm 0.0026$ \\
				\midrule  
				\multirow{3}{*}{$\dfrac{s}{2}$}
				& Euc &$0.953\pm0.0030$ &$\textbf{0.923}\pm0.0048$ &$\textbf{0.941}\pm0.0027$ &$0.905\pm0.0033$ &$\textbf{0.901}\pm0.0038$ &$\textbf{0.903}\pm0.0024$ \\
				& Cos & $\textbf{0.959} \pm 0.0198$ & $0.448 \pm 0.0997$ & $0.719 \pm 0.0411$ & $0.903 \pm 0.0034$ & $0.446 \pm 0.0042$ & $0.690 \pm 0.0025$ \\
				& Mah & $0.941 \pm 0.0033$ & $0.895 \pm 0.0046$ & $0.920 \pm 0.0026$ & $0.898 \pm 0.0034$ & $\underline{0.825} \pm 0.0046$ & $\underline{0.866} \pm 0.0027$ \\
				\bottomrule
			\end{tabular}%
		}
	\end{table}

The interaction scale $\sigma$ plays a critical role in identifying new categories. We let $\sigma=1$ be the initial value in the iteration. It is also the upper bound of $\sigma$. When an input sample activates no high-level neuron, we assign it to a new category. To verify the rationality in this strategy, we allow the $\sigma$ to increase by $\sigma/\lambda$ for 0, 1, 4, and 9 times and compare the results, as shown in Table \ref{aresult3} and \ref{aresult4}. We can see that the proposed model achieves the highest accuracy with fixed $\sigma=1$ as the upper bound. Its performance decreases with increasing upper bound of $\sigma$.

	\begin{table}[h]
		\caption{ Results on CIFAR-10, CIFAR100, and ImageNet-100 with fixed and increasing upper bound of $\sigma$.}
		\label{aresult3}
		\centering
		\resizebox{\linewidth}{!}{%
			\begin{tabular}{c|ccc|ccc|ccc}
				\toprule
				\multirow{2}{*}{\textbf{Times}} & \multicolumn{3}{c|}{\textbf{CIFAR-10}} & \multicolumn{3}{c|}{\textbf{CIFAR-100}} & \multicolumn{3}{c}{\textbf{ImageNet-100}} \\
				\cline{2-10}
				& Old& New & All& Old& New & All & Old& New & All \\
				\midrule  
				0 &$\textbf{0.990}\pm0.0007$  &$\textbf{0.975}\pm0.0023$  &$\textbf{0.983}\pm0.0011$  &$\textbf{0.993}\pm0.0006$  &$\textbf{0.978}\pm0.0019$  &$\textbf{0.986}\pm0.0009$  &$\textbf{0.998}\pm0.0003$  &$\textbf{0.986}\pm0.0013$  &$\textbf{0.992}\pm0.0006$ \\
				1 &$\underline{0.983}\pm0.0008$  &$\underline{0.937}\pm0.0030$  &$\underline{0.962}\pm0.0014$  &$\underline{0.988}\pm0.0008$  &$\underline{0.957}\pm0.002$  &$\underline{0.973}\pm0.0011$  &$\underline{0.9958}\pm0.0004$  &$\underline{0.981}\pm0.0016$  &$\underline{0.982}\pm0.0008$  \\
				4 &$0.980\pm0.0009$  &$0.598\pm0.0095$  &$0.793\pm0.0047$  &$0.985\pm0.0008$  &$0.582\pm0.0092$  &$0.788\pm0.0046$  &$0.994\pm0.0005$&$0.491\pm0.0084$&$0.746\pm0.0042$\\
				9&$0.980\pm0.0010$&$0.230\pm0.0045$ &$0.606\pm0.0023$  &$0.985\pm0.0009$  &$0.228\pm0.0043$&$0.607\pm0.0022$  &$0.994\pm0.0005$  &$0.231\pm0.0043$ &$0.613\pm0.0022$\\
				\bottomrule
			\end{tabular}%
		}
	\end{table}
	
	\begin{table}[h]
		\caption{Results on CUB-200 and Stanford Cars with fixed and increasing upper bound of $\sigma$.}
		\label{aresult4}
		\resizebox{\linewidth}{!}{%
			\centering
			\begin{tabular}{c|ccc|ccc}
				\toprule
				\multirow{2}{*}{\textbf{Times}} & \multicolumn{3}{c|}{\textbf{CUB-200}} & \multicolumn{3}{c}{\textbf{Stanford Cars}} \\
				\cline{2-7}
				& Old& New & All& Old& New & All\\
				\midrule  
				0  &$\textbf{0.953}\pm0.0030$ &$\textbf{0.923}\pm0.0048$ &$\textbf{0.941}\pm0.0027$ &$\textbf{0.905}\pm0.0033$ &$\textbf{0.901}\pm0.0038$ &$\textbf{0.903}\pm0.0024$ \\
				1  &$\underline{0.942}\pm0.0034$ &$\underline{0.890}\pm0.0045$ &$\underline{0.935}\pm0.0028$ &$\underline{0.895}\pm0.0034$ &$\underline{0.820}\pm0.0048$ &$\underline{0.862}\pm0.0025$\\
				4  &$0.934\pm0.0035$ &$0.510\pm0.0087$ &$0.724\pm0.0040$  &$0.882\pm0.0037$ &$0.592\pm0.0074$ &$0.742\pm0.0037$ \\
				9 &$0.929\pm0.0038$ &$0.254\pm0.0050$  &$0.586\pm0.0026$  &$0.883\pm0.0035$&$0.215\pm0.0046$ &$0.550\pm0.0028$\\
				\bottomrule
			\end{tabular}%
		}
	\end{table}

To evaluate the impact of selection of the iteration ratio $\lambda$, we test the proposed model with $\lambda=0.2, 0.4, 0.5, 0.6, 0.8$, as shown in Table \ref{aresult5} and \ref{aresult6}. 

ViTNF achieves peak accuracy and balance between Old and New classes at $\lambda = 0.4$. It yields the best overall performance across all the datasets. For CIFAR-10, it achieves an accuracy of 98.3\% for all classes, with 97.5\% for new classes. On CIFAR-100, it achieves 98.6\% for all classes and 97.8\% for new classes. On ImageNet-100, it achieves 99.2\% for all classes and 98.6\% for new classes. On CUB-200, it achieves 94.1\% for all classes and 92.3\% for new classes. On Stanford Cars, it achieves 90.3\% for all classes and 90.1\% for new classes. It sustains a balanced performance on new and old classes with a slight decline of no more than +3\%. Notably, on Stanford Cars, $\lambda=0.4$ achieves 90.1\% new accuracy—nearly matching old class accuracy (90.5\%), indicating unbiased generalization.

For the other values, $\lambda = 0.2$ achieves high accuracy for old classes but struggles with new classes (92.0\% on CIFAR-100), indicating an imbalanced performance. $\lambda = 0.5$ provides a close second to $\lambda=0.4$, but with marginally lower new class accuracy (e.g., 97.4\% vs. 97.8\% on CIFAR-100). Higher values of $\lambda$ lead to a gradual decline in accuracy for both old and new classes (e.g., CUB-200 All drops from 94.1\% ($\lambda=0.4$) to 91.7\% ($\lambda=0.8$)).

The experiments conclusively identify $\lambda=0.4$ as the optimal iteration ratio for ViTNF, delivering state-of-the-art accuracy, stability, and balance between the old and new classes. This finding validates the design choice for iterative refinement in GCD tasks.

	\begin{table}[h]
		\caption{Results on CIFAR-10, CIFAR100, ImageNet-100 with different $\lambda$.}
		\label{aresult5}
		\centering
		\resizebox{\linewidth}{!}{%
			\begin{tabular}{c|ccc|ccc|ccc}
				\toprule
				\multirow{2}{*}{\textbf{$\lambda$}} & \multicolumn{3}{c|}{\textbf{CIFAR-10}} & \multicolumn{3}{c|}{\textbf{CIFAR-100}} & \multicolumn{3}{c}{\textbf{ImageNet-100}} \\
				\cline{2-10}
				& Old& New & All& Old& New & All & Old& New & All \\
				\midrule  
				0.2 & $\textbf{0.993}\pm0.0006$ & $0.920\pm0.0060$ & $0.960\pm0.0027$ & $\textbf{0.995}\pm0.0005$ & $0.922\pm0.0051$ & $0.960\pm0.0024$&$\textbf{0.999}\pm0.0001$&$0.911\pm0.0054$&$0.957\pm0.0026$ \\
				0.4 &$\underline{0.990}\pm0.0007$  &$\textbf{0.975}\pm0.0023$  &$\textbf{0.983}\pm0.0011$  &$\underline{0.993}\pm0.0006$  &$\textbf{0.978}\pm0.0019$  &$\textbf{0.986}\pm0.0009$  &$\underline{0.998}\pm0.0003$  &$\textbf{0.986}\pm0.0013$  &$\textbf{0.992}\pm0.0006$  \\
				0.5 &$0.988\pm0.0007$  &$\underline{0.974}\pm0.0019$  &$\underline{0.982}\pm0.0009$  &$0.991\pm0.0007$  &$\textbf{0.978}\pm0.0017$  &$\underline{0.985}\pm0.0009$  &$0.997\pm0.0004$&$\underline{0.983}\pm0.0016$&$\underline{0.990}\pm0.0008$\\
				0.6&$0.986\pm0.0008$&$0.967\pm0.0022$ &$0.977\pm0.0011$  &$0.990\pm0.0007$  &$\underline{0.972}\pm0.0020$&$0.981\pm0.0010$  &$0.996\pm0.0004$  &$0.981\pm0.0015$ &$0.989\pm0.0007$\\
				0.8&$0.983\pm0.0009$&$0.949\pm0.0028$ &$0.967\pm0.0014$  &$0.988\pm0.0008$  &$0.961\pm0.0021$&$0.975\pm0.0011$  &$0.995\pm0.0005$  &$0.969\pm0.0018$ &$0.983\pm0.0009$\\
				\bottomrule
			\end{tabular}%
		}
	\end{table}
	\begin{table}[h]
		\caption{Results on CUB-200 and Stanford Cars with different $\lambda$}
		\label{aresult6}
		\resizebox{\linewidth}{!}{%
			\centering
			\begin{tabular}{c|ccc|ccc}
				\toprule
				\multirow{2}{*}{\textbf{$\lambda$}} & \multicolumn{3}{c|}{\textbf{CUB-200}} & \multicolumn{3}{c}{\textbf{Stanford Cars}} \\
				\cline{2-7}
				& Old& New & All& Old& New & All\\
				\midrule  
				0.2  &$\textbf{0.965}\pm0.0024$&$0.856\pm0.0070$ &$0.914\pm0.0036$ &$\textbf{0.906}\pm0.0035$  &$\underline{0.893}\pm0.0052$  &$\textbf{0.903}\pm0.0029$ \\
				0.4  &$\underline{0.953}\pm0.0030$ &$\underline{0.923}\pm0.0048$ &$\textbf{0.941}\pm0.0027$ &$\underline{0.905}\pm0.0033$ &$\textbf{0.901}\pm0.0038$ &$\textbf{0.903}\pm0.0024$\\
				0.5  &$0.945\pm0.0032$ &$\textbf{0.927}\pm0.0043$ &$\underline{0.938}\pm0.0025$  &$0.904\pm0.0033$ &$0.884\pm0.0041$ &$\underline{0.898}\pm0.0025$ \\
				0.6 &$0.945\pm0.0032$ &$\underline{0.923}\pm0.0049$  &$0.935\pm0.0028$  &$0.900\pm0.0033$&$0.866\pm0.0042$ &$0.887\pm0.0025$\\
				0.8 &$0.936\pm0.0035$ &$0.896\pm0.0047$  &$0.917\pm0.0028$  &$0.858\pm0.0036$&$0.836\pm0.0047$ &$0.864\pm0.0028$\\
				\bottomrule
			\end{tabular}%
		}
	\end{table}

	\section{Conclusion}\label{conclusion}
In this paper, we present a novel architecture for generalized category discovery (GCD) by combining the feature extractor of ViT with a neural field-based classifier. We first present a new static neural field function to describe the activity distribution of the neural field and then use two static neural field functions to build an efficient few-shot classifier. By replacing the MLP head responsible for classification in ViT with our proposed NF classifier, we propose an effective few-shot learning model ViTNF with powerful GCD capability. Extensive experiments demonstrate the effectiveness of ViTNF. It achieves far superior accuracy to existing state-of-the-art algorithms on the CIFAR-10, CIFAR-100, ImageNet-100, CUB-200, and Stanford Cars datasets without using meta-training and fine-tuning.

In future work, we plan to further explore the potential of our proposed method by applying ViTNF to other tasks such as medical image classification and object detection.
	
%	\section{Limitation}\label{limitation}
%	In this paper, we have proposed a few-shot learning method for the GCD task. As the experimental results above, our method demonstrates certain advantages in the GCD task. However, there is still a gap between our results and the best results in some specific datasets. Additionally, as new-class images are learned, the model's ability to learn and recognize old-class images may be impacted to some extent. We hope to make improvements and enhancements in these areas in future work.\par
%	
	
	\bibliographystyle{plain} % 或者其他样式，如 apalike、plainnat 等
	\bibliography{ref} % 引用 ref.bib 文件

\end{document}